%% file: main.tex
\documentclass[conference]{IEEEtran}
\IEEEoverridecommandlockouts
\usepackage{amsmath,amssymb,amsfonts}
\usepackage{algorithmic}
\usepackage{graphicx}
\usepackage{textcomp}
\usepackage[table]{xcolor}
\usepackage{ragged2e}
\usepackage{booktabs}
\usepackage[flushleft]{threeparttable}
\usepackage{array}
\usepackage{multirow} 
\usepackage{hyperref}
\def\BibTeX{{\rm B\kern-.05em{\sc i\kern-.025em b}\kern-.08em
    T\kern-.1667em\lower.7ex\hbox{E}\kern-.125emX}}
\begin{document}
\title{ChiseLLM: Unleashing the Power of Reasoning LLMs for Chisel Agile Hardware Development}

\makeatletter
\newcommand{\linebreakand}{%
    \end{@IEEEauthorhalign}
    \hfill\mbox{}\par
    \mbox{}\hfill\begin{@IEEEauthorhalign}
}
\makeatother

\author{\IEEEauthorblockN{Bowei Wang}
    \IEEEauthorblockA{\textit{Computer Department} \\
        \textit{National University of Defense Technology}\\
        Changsha, China \\
        wangbowei@nudt.edu.cn}
    \and
    \IEEEauthorblockN{Jiaran Gao}
    \IEEEauthorblockA{\textit{School of Computer \& Communication Engineering} \\
        \textit{University of Science and Technology Beijing}\\
        Beijing, China \\
        U202241904@xs.ustb.edu.cn}
    \linebreakand
    \IEEEauthorblockN{Yelai Feng}
    \IEEEauthorblockA{\textit{Intelligent Microelectronics Center} \\
        \textit{Qiyuan Lab}\\
        Beijing, China \\
        fengyelai@qiyuanlab.com}
    \and
    \IEEEauthorblockN{Renzhi Chen}
    \IEEEauthorblockA{\textit{Intelligent Microelectronics Center} \\
        \textit{Qiyuan Lab}\\
        Beijing, China \\
        chenrenzhi@qiyuanlab.com}
    \linebreakand
    \IEEEauthorblockN{Shanshan Li\textsuperscript{*}}
    \IEEEauthorblockA{\textit{College of Computer Science and Technology} \\
        \textit{National University of Defense Technology}\\
        Changsha, China \\
        shanshanli@nudt.edu.cn}
    \and
    \IEEEauthorblockN{Lei Wang\textsuperscript{*}}
    \IEEEauthorblockA{\textit{Defense Innovation Institute} \\
        \textit{Academy of Military Science}\\
        Beijing, China \\
        leiwang@nudt.edu.cn}
    \thanks{*Corresponding authors.}
}

\maketitle

\input{sections/abstract}

\begin{IEEEkeywords}
    Large Language Models, Chisel Language, Agile Hardware Development
\end{IEEEkeywords}

\definecolor{green}{RGB}{180,255,180}
\definecolor{red}{RGB}{255,225,225}
\definecolor{blue}{RGB}{180,180,255}

\input{sections/introduction}
\input{sections/related-work}
\input{sections/preliminary}
\input{sections/method}

\input{sections/experiments}
\input{sections/case-study}
\input{sections/conclusion}


\bibliographystyle{IEEEtran}
\bibliography{main}

\end{document}

%% file: sections/abstract.tex
\begin{abstract}
    The growing demand for Domain-Specific Architecture (DSA) has driven the development of Agile Hardware Development Methodology (AHDM). Hardware Construction Language (HCL) like Chisel offers high-level abstraction features, making it an ideal language for HCL-Based AHDM. While Large Language Models (LLMs) excel in code generation tasks, they still face challenges with Chisel generation, particularly regarding syntax correctness and design variability. Recent reasoning models have significantly enhanced code generation capabilities through test-time scaling techniques. However, we found that reasoning models without domain adaptation cannot bring substantial benefits to Chisel code generation tasks. This paper presents ChiseLLM, a solution comprising data processing and transformation, prompt-guided reasoning trace synthesis, and domain-adapted model training. We constructed high-quality datasets from public RTL code resources and guided the model to adopt structured thinking patterns through prompt enhancement methods. Experiments demonstrate that our ChiseLLM-7B and ChiseLLM-32B models improved syntax correctness by 18.85\% and 26.32\% respectively over base models, while increasing variability design ability by 47.58\% compared to baseline reasoning models. Our datasets and models are publicly available, providing high-performance, cost-effective models for HCL-Based AHDM, and offering an effective baseline for future research. Github repository: \url{https://github.com/observerw/ChiseLLM}
\end{abstract}

%% file: sections/introduction.tex
\section{Introduction}
\label{sec:introduction}

\input{figures/introduction}

With the rapid advancement of artificial intelligence and big data, the market demand for Domain-Specific Architecture (DSA) has grown significantly. This trend has promoted the development of Agile Hardware Development Methodology (AHDM)~\cite{lee2016agileriscv,bahrCreatingAgileHardware2020}. AHDM imposes higher requirements on shortening hardware design cycles and enhancing design variability, which refers to the adaptability of hardware design to changes in requirements or constraints~\cite{acherProgrammingVariabilityLarge2023,brinkHardwareVariabilityRelation2014} (see Section~\ref{sec:preliminary}). To address these requirements, next-generation Hardware Construction Languages (HCL)~\cite{bachrachChiselConstructingHardware2012,paponSpinalHDL2025}, represented by Chisel, integrate modern programming language features into hardware design, significantly enhancing the abstract expression capability of hardware design. Consequently, HCL-Based AHDM shows promising prospects and has been applied in several large-scale projects~\cite{micro2022xiangshan,ChipsallianceRocketchip2025}.

Large Language Models (LLMs) have demonstrated impressive code generation capabilities~\cite{jain2024livecodebench}, prompting researchers to explore the ability of using LLMs for AHDM development~\cite{pinckneyRevisitingVerilogEvalYear2025}, including automated module-level design~\cite{liu2024rtlcodera} and architectural design~\cite{chang2023chipgpt}. The application of LLMs for HCL-Based AHDM development has gained preliminary research attention~\cite{niu2025rechisel,liuChatChiselEnablingAgile2024}, indicating that this approach can more effectively manage project complexity, reduce migration costs, and enhance the practicality of agile methods. However, we have observed that existing smaller-scale open-source models lack sufficient capability in Chisel code generation: models frequently produce syntax/semantic errors when generating Chisel code and struggle to leverage Chisel's language features for effective variability design. This limitation hinders the potential of the generated code in tasks such as parameterized module generation or design space exploration. These challenges impede the practical application of LLMs in HCL-Based AHDM development.

Recent reasoning models represented by OpenAI-o1~\cite{IntroducingOpenAIO12024} and Deepseek-R1~\cite{deepseekaiDeepSeekR1IncentivizingReasoning2025} introduce test-time scaling~\cite{snell2024scaling,welleck2024decoding}, conducting deliberate thinking and reflection before generating final results. Related studies show that reasoning models can generate high-quality code more effectively than non-reasoning models that lack thinking processes~\cite{jain2024livecodebench,O1TopsAiders2024,li2025stesttimescaling}. Therefore, we consider the introduction of test-time scaling as a promising approach to enhance model performance in Chisel code generation tasks. Nevertheless, our experiments indicate that \textbf{reasoning models without domain adaptation cannot achieve significant additional benefits in Chisel code generation scenarios} (see Section~\ref{sec:experiment-rq1}), as general reasoning models lack reliable hardware logic thinking patterns, leading to outputs prone to hallucinations. Consequently, \textit{adapting reasoning models to reason with task-specific thinking patterns} becomes a key challenge.

To address this challenge, we propose the \textbf{ChiseLLM} series of datasets and models. Our approach includes: (1) Original dataset processing and transformation: we collected Chisel and Verilog code from publicly available RTL code sources, which after processing and transformation formed two high-quality domain-specific instruction fine-tuning datasets; (2) Prompt-guided reasoning trace data synthesis: we used a prompt-guided approach to construct reasoning datasets ChiseLLM-\{Completion,Decompile\} with specific thinking patterns, enabling models to learn how to reason with these specific thinking patterns; (3) Reasoning model training: based on the generated data, we fine-tuned the Qwen2.5-Coder~\cite{hui2024qwen25coder} series to produce the ChiseLLM-\{7,32\}B models. Experimental results demonstrate that \textbf{ChiseLLM successfully adapts reasoning models to achieve excellent Chisel code generation capabilities}. In terms of syntactic/semantic correctness, the ChiseLLM-7B model improved by an average of 18.85\% in syntactic correctness compared to the pre-fine-tuned base model, while the ChiseLLM-32B model improved by an average of 26.32\% in Pass@5 compared to the base model, with an average improvement of 20.44\% over the optimal baseline across all tasks, achieving performance comparable to commercial models in syntactic correctness and decompilation accuracy. Regarding variability capability, our model showed a 47.58\% improvement compared to the baseline reasoning model, indicating that the ChiseLLM model can effectively generate designs with high variability.

Our main contributions are as follows:

\begin{itemize}
    \item We introduce the ChiseLLM series of models and datasets, successfully elevating the performance of open-source LLMs in Chisel code generation tasks to a practical and effective level, and achieving performance comparable to commercial models in certain tasks, while controlling the model size to below 32B to reduce computational costs and provide broader applicability.
    \item Our prompt-guided method successfully adapts reasoning models for domain-specific use, introducing new approaches for LLM-assisted hardware design. We have constructed a standardized evaluation system to comprehensively assess the performance of mainstream code generation models on Chisel tasks.
    \item Our models, datasets, and construction process will be open-sourced for use and improvement by subsequent researchers. We hope that ChiseLLM can serve as a foundational model in the HCL-Based AHDM field, driving the advancement of related research.
\end{itemize}

%% file: figures/introduction.tex
\begin{figure*}
    \centering
    \includegraphics[width=0.95\textwidth]{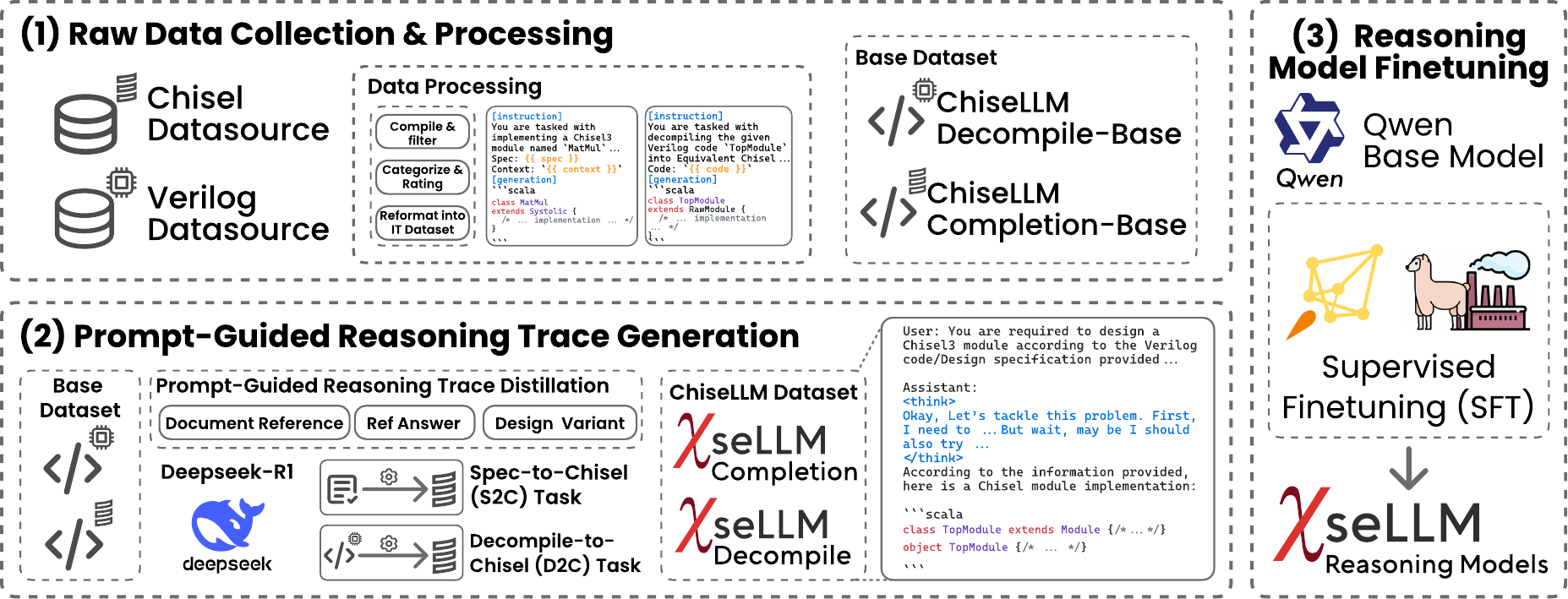}
    \caption{An overview diagram of the construction of the ChiseLLM datasets and models, including Source Data Processing \& Synthesing, Prompt-Guided Reasoning Trace Generation and Reasoning Model Finetuning.}
    \label{fig:intro-overview}
\end{figure*}

%% file: sections/related-work.tex
\section{Related Work}
\label{sec:related-work}

\subsection{Hardware Construction Languages}
\label{sec:related-work-agile-hardware}

To meet the primary goals of agile hardware development of enhancing design flexibility and reducing development cycles, diverse approaches have been explored, particularly the concept of "hardware generators" that can automatically produce designs tailored to specific requirements. Before the emergence of HCL, projects such as Genesis2~\cite{solomatnikovChipMultiProcessor2007} attempted to address this need by treating hardware code as templates and automatically generating hardware designs through embedded language. The evolution of this approach led to the development of HCL, which integrate generation logic into high-level programming languages. HCLs are defined as Domain Specific Languages (DSLs) designed on top of high-level languages, such as Chisel on Scala and PyRTL on Python~\cite{clow2017pythonic}. The original design intention of HCLs was to standardize and formalize these design generators, utilizing the expressive capabilities of host languages to automatically generate designs based on a set of high-level design parameters and constraints~\cite{MotivationChisel}, thereby eliminating the need to reimplement designs for each class of design specifications. This approach reduces the cost of developing new languages from scratch while introducing important concepts and mature technologies from software engineering into hardware design. Currently, the Chisel language has been employed in multiple large-scale agile hardware development projects, including Rocket-Chip~\cite{ChipsallianceRocketchip2025}, XiangShan~\cite{micro2022xiangshan}, Berkeley BOOM Processor~\cite{RiscvboomRiscvboomSonicBOOM}, etc.

\subsection{Reasoning Models in Code Generation}
\label{sec:related-work-reasoning-model-codegen}

Test-Time Scaling techniques refer to post-training modifications~\cite{kumar2025llm} of large language models, enabling them to conduct extended Chain-of-Thought~\cite{weiChainofthoughtPromptingElicits2023} reasoning processes (known as reasoning traces), thereby achieving better performance in tasks requiring reasoning (such as mathematics and code generation). Models with such capabilities are termed Reasoning Models, including OpenAI-o series~\cite{IntroducingOpenAIO12024,IntroducingOpenAIO3}, Deepseek-R1~\cite{deepseekaiDeepSeekR1IncentivizingReasoning2025}, Claude-3.7 Sonnet~\cite{Claude37Sonnet}, Gemini 2.5~\cite{Gemini25Our2025}, and others. Reasoning models have demonstrated powerful performance in code generation tasks~\cite{jain2024livecodebench,IntroducingSWEbenchVerified}, making them a significant research direction in this field. Nevertheless, LLM performance remains limited in domain-specific languages (DSLs) with scarce corpus~\cite{gu2025effectiveness}, which is widely acknowledged to require Domain Adaptation~\cite{joelSurveyLLMbasedCode2024}. Our work confirms this conclusion, indicating that reasoning models still require domain adaptation to achieve optimal performance in Chisel code generation tasks.

%% file: sections/preliminary.tex
\section{Preliminary}
\label{sec:preliminary}

\subsection{Agile Hardware Development and Design Variability}

AHDM draws inspiration from the software agile development concept \cite{ManifestoAgileSoftware}, aiming to enhance the flexibility and responsiveness of hardware design through rapid iterations and continuous feedback. Consequently, projects adopting AHDM should possess a high degree of variability, which is defined as the ability of a project to adapt to changes in market demands or contextual conditions \cite{cortellessaFundamentalApproachesSoftware2013}. In the hardware domain, variability primarily emphasizes the adaptability of hardware designs to requirement or constraint changes, thereby supporting the need for rapid iterations. This paper utilizes the following direct inference: \textit{highly variable hardware designs should be able to meet different design requirement changes with minimal code refactoring costs}. For instance, variability design can satisfy different functional and performance constraints by merely modifying a set of high-level parameters without requiring extensive modifications to the underlying implementation, as exemplified by \cite{ChipsallianceRocketchip2025}.

\subsection{Chisel Generation Tasks}

Current research on using LLMs for AHDM development primarily focuses on transforming design specifications into traditional HDL code. However, Chisel serves as an intermediate layer between design specifications and traditional HDL, necessitating consideration of two distinct scenarios for code generation: directly generating Chisel code from design specifications (Spec-to-Chisel) and converting existing traditional HDL code into equivalent Chisel code (Decompile-to-Chisel). This classification reflects two practical engineering needs: development of new designs and upgrading of existing ones. Table \ref{tab:preliminary-tasks} summarizes these two types of tasks.

\begin{table}[!t]
    \caption{Task types and descriptions in Chisel code generation.}
    \label{tab:preliminary-tasks}
    \begin{center}
        \begin{tabular}{>{\centering}p{2.5cm}>{\centering}p{3cm}>{\centering\arraybackslash}p{1.8cm}}
            \toprule
            \textbf{Task Name}                 & \textbf{Description}                                                                & \textbf{Reference}                                        \\
            \midrule
            \textbf{Spec-to-Chisel (S2C)}      & Generate Chisel code that meets functional requirements from design specifications  & \cite{liuChatChiselEnablingAgile2024}                     \\
            \midrule
            \textbf{Decompile-to-Chisel (D2C)} & Convert low-level RTL code to Chisel that at least satisfies the same functionality & \cite{bruant2020sv2chisel,bruantAgileHardwareDesigns2021} \\
            \bottomrule
        \end{tabular}
    \end{center}
\end{table}

%% file: sections/method.tex
\section{Method}
\label{sec:method}

In this section, we introduce the raw data collection and processing (Section~\ref{sec:method-dataset-preprocess}), prompt-guided distillation (Section~\ref{sec:method-prompt-guided-distillation}), and model training (Section~\ref{sec:method-finetuning}) process, which together form the complete data construction and model training pipeline for ChiseLLM.

\subsection{Raw Dataset Collection and Preprocessing}
\label{sec:method-dataset-preprocess}

Publicly available HDL code data is extremely scarce. (System)Verilog accounts for only $3.2 \times 10^{-2}\%$ of the the-stack-v2-dedup dataset, while HCLs like Chisel have an even lower representation (approximately $1 \times 10^{-3}\%$). To our knowledge, there is currently no publicly available large-scale training dataset specifically for the Chisel language. To address this limitation, we leveraged Chisel's characteristics as an intermediate layer language and utilized both Chisel and Verilog data to expand the size and diversity of our dataset. We collected publicly accessible Chisel and Verilog datasets, processed and transformed them separately, resulting in two high-quality foundational datasets: ChiseLLM-\{Completion,Decompile\}-Base. These two datasets contain Chisel code completion and code decompilation tasks, respectively.


ChiseLLM-Completion-Base is derived from the the-stack-v2-dedup dataset \cite{lozhkov2024starcoder}, which is a large pre-training dataset containing multiple programming languages. We extracted Chisel3 code files from this dataset, excluding any files containing \texttt{chiseltest} or \texttt{scalatest} to remove all test code. This filtering was necessary due to the instability of the Chisel language testing framework \cite{UcbbarChiseltest2025} and its misalignment with our task objectives. After filtering and length screening, we obtained approximately 8.3K Chisel code files. Subsequently, we utilized large language models to add detailed comments to the code and converted the complete code into question-answer pairs for code completion tasks, which included design specifications and contextual information.

\subsection{Prompt-Guided Distillation}
\label{sec:method-prompt-guided-distillation}

We followed the methodology proposed in the Deepseek-R1 paper \cite{deepseekaiDeepSeekR1IncentivizingReasoning2025}, using Deepseek to generate reasoning traces for the instruction fine-tuning dataset, thereby creating a reasoning dataset to train smaller models to acquire reasoning capabilities. This process is known as distillation. Related research generally suggests that LLMs primarily learn how to effectively utilize knowledge acquired during pre-training when fine-tuning rather than learning new knowledge \cite{zhou2023lima}. Based on this insight, we focused on constructing reasoning data with specific "thinking patterns". As shown in Figure~\ref{fig:method}(b), to obtain reasoning data with the desired thinking patterns, we employed a prompt-guided approach, providing Deepseek-R1 with additional task-related prompts and guiding information to direct the model toward generating reasoning traces more aligned with task requirements. We refer to this process as prompt-guided distillation. For both Spec-to-Chisel and Decompile-to-Chisel tasks, we adopted different prompt-guided approaches tailored to each task objective. By learning from these data, models can adopt appropriate thinking patterns when facing specific tasks, demonstrating better problem-solving capabilities than generic thinking patterns. Through this method, we processed the ChiseLLM-\{Completion,Decompile\}-Base datasets, ultimately creating two high-quality reasoning datasets: ChiseLLM-\{Completion,Decompile\}.


\input{figures/method}

\subsubsection{Spec-to-Chisel Task}
\label{sec:method-distillation-spec2chisel}

The objective of the Spec-to-Chisel task is to convert design specifications into Chisel code. We observed that when generating code, models typically only reason about functional implementation while lacking reflection on syntax. This leads to more frequent hallucinations, including references to non-existent Chisel APIs or incorrect use of syntax from other languages. To enable models to reasonably utilize Chisel language features in their reasoning traces, we provided two types of prompting guidance: document references and benchmark answers. The expected thinking pattern is: during the reasoning process, the model first forms logical steps for functional implementation, then reflects on the syntax level (achieved through "recollection" of documentation), ultimately generating syntactically and functionally correct answers.

\textbf{Document References.} We downloaded all language documentation from the Chisel official website and processed it into independently citable Markdown document fragments. We created an index document for the official documentation, including titles and content summaries for all document fragments, assigning each a unique chapter ID. Subsequently, we provided this document as context, having the Qwen2.5-72B-Instruct annotate the code at the line level in the form of comments. Afterward, document fragments were matched according to their chapter IDs and included as part of the corresponding information for the code data. After processing, each code corresponded to several relevant documents, averaging between 5-10 documents. We included these documents as part of the contextual information during distillation, requiring the model to utilize this documentation when reflecting on syntax.

\textbf{Benchmark Answers.} The current common method for constructing reasoning datasets involves rejection sampling of independently verifiable, high-quality code datasets \cite{penedo2025codeforces, OpenThoughts}. However, existing Chisel code data exists in the form of pre-training datasets, lacking high-quality reference answers and test cases. Generated Chisel code is difficult to judge for correctness due to the absence of unit test cases and compilation challenges. To address this, we included benchmark answers as part of the guidance, allowing the reasoning model to generate intermediate reasoning traces when given reference answers. This approach prevented the model's reasoning traces from leading to incorrect answers.

\subsubsection{Decompile-to-Chisel Task}
\label{sec:method-distillation-decompile2chisel}

The objective of the Decompile-to-Chisel task is to convert low-level HDL code into Chisel code. Existing work \cite{bruant2020sv2chisel} can programmatically convert Verilog code to Chisel code, but it can only achieve unoptimized mapping of code structures, with post-mapping optimization still requiring manual effort. We believe that LLMs have greater potential in this task, capable of achieving high-quality conversion and optimization through understanding and reasoning about the code. To this end, we provided two types of prompting guidance: variant pattern specifications and Chisel feature descriptions. Our expected thinking pattern is: \textit{during the reasoning process, the model can understand the core functionality of the original code, infer potential design requirement changes by applying variant pattern to the original code, and utilize Chisel language features to restructure the code to cover all variants.}

\textbf{Variant Pattern Specifications.} Based on the characteristics of variability design (see Section~\ref{sec:preliminary}), we required the model to first consider possible requirement changes the module might face, then optimize the code accordingly to reduce the refactoring cost of adapting to each changes. To guide the model in producing reasonable design variants, we constrained the form of variant into the following three types:

\input{figures/variation}

\begin{itemize}
    \item \textbf{Configurable Variant:} A modified version that extracts adjustable attributes (such as data width, cache depth) as configurable parameters without changing core functionality or interfaces. This type of variant is very common in Design Space Exploration, typically used for performance optimization under different design requirements.
    \item \textbf{Functional Variant:} A modified version that changes input-output behavior (such as protocol version support, exception handling mechanisms, algorithm implementation), directly modifying the functional requirements of the design. This type of variant typically appears during rapid prototype design in early design stages or during functional expansion in later stages.
    \item \textbf{Structural Variant:} A modified version that adjusts hardware architecture (such as pipeline stages, memory organization, FSM encoding schemes) while preserving functional equivalence, affecting timing/area. This type of variant typically appears after functional requirements have been determined, during PPA (Power, Performance, Area) optimization.
\end{itemize}

We required the model to provide configurable variant, and only one of either functional or structural variants to avoid guiding toward overly complex and unnecessary designs. Subsequently, we required the model to design a Chisel module that can cover all variants.

\textbf{Chisel Feature Descriptions.} To enable the model to reasonably utilize Chisel language features when thinking, we provided a brief description of Chisel's language features and design patterns, including parameterization, functional components and functional programming, combinational logic generation, generic parameters, optional interfaces, inheritance, and traits. We required the model to select appropriate language features based on the selected variant pattern and the characteristics of the problem itself when generating reasoning traces. For example, when transforming an up-counter module into a down-counter module as a functional variant, the model should first abstract the counter behavior as \texttt{abstract class CounterBase}, then implement the corresponding functionality through implementation of the abstract interface.

\subsection{Model Finetuning}
\label{sec:method-finetuning}

We used LLaMA-Factory \cite{zheng2024llamafactory} for full-parameter fine-tuning of our models. We first analyzed the data characteristics of the ChiseLLM-Completion and ChiseLLM-Decompile datasets, as shown in Figure~\ref{fig:method}(c). The token length of the Decompile dataset is significantly longer than that of the Completion dataset, with an average around 9K tokens. Given the difficulty characteristics of both task types, this statistical distribution is expected. Therefore, we mixed the data according to proportion, with a sampling ratio of 3:7, and trained on both Qwen2.5-7B-Instruct and Qwen2.5-32B-Instruct, ultimately obtaining a series of models with reasoning capabilities, called the ChiseLLM Models. All models are publicly available.


%% file: figures/method.tex
\begin{figure}[t!]
    \centering
    \includegraphics[width=\linewidth]{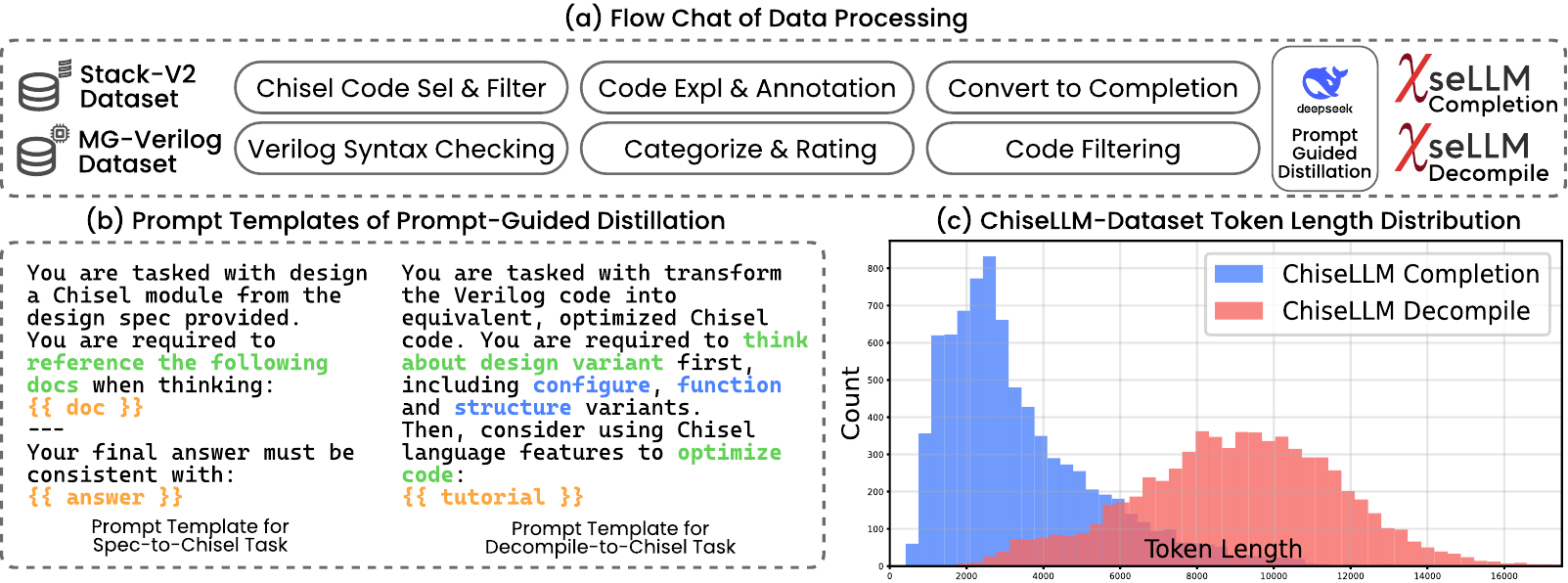}
    \caption{Diagrams related to the ChiseLLM data processing and distillation workflow. Figure (a) illustrate the process of collecting and processing the source data. Figure (b) includes the prompt template used for prompt-guided distillation. Figure (c) shows the statistical characteristics of the ChiseLLM-\{Completion,Decompile\} datasets.}
    \label{fig:method}
\end{figure}

%% file: figures/variation.tex
\begin{figure}[t!]
    \centering
    \includegraphics[width=0.95\linewidth]{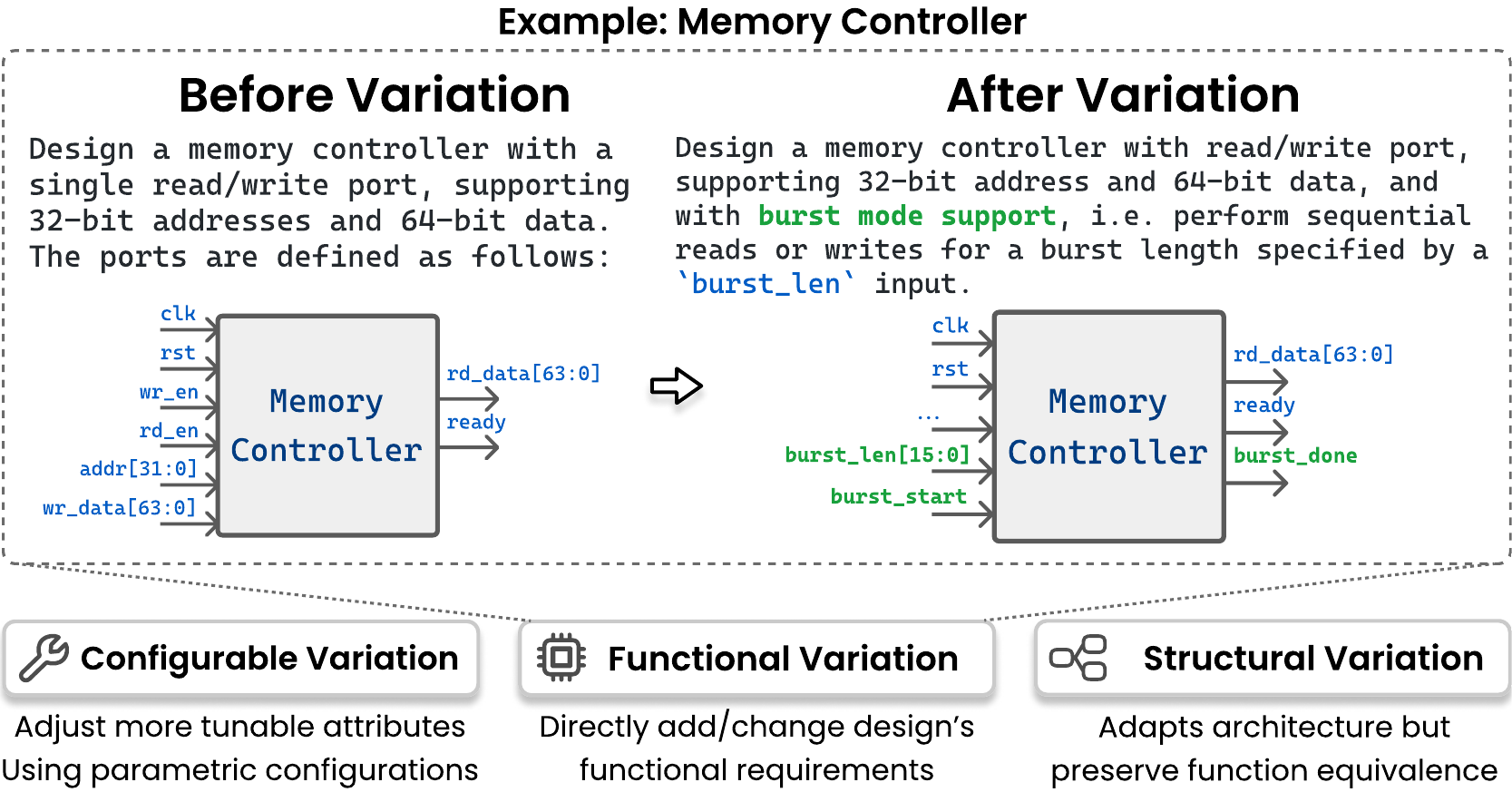}
    \caption{Schematic diagrams of the three types of design variant. Practical examples of functional variants are shown in detail in the figure.}
    \label{fig:method-variation}
\end{figure}

%% file: sections/experiments.tex
\section{Experiments}
\label{sec:experiments}

\input{tables/rq1}

In this section, we conduct a comprehensive evaluation of ChiseLLM models. We assess the performance of ChiseLLM models and baseline models on Spec-to-Chisel and Decompile-to-Chisel tasks, and analyze the capability of models in variability design. We also conduct ablation studies to analyze the impact of ChiseLLM datasets on model performance.

\subsection{Research Questions}
\label{sec:experiment-research-questions}

Our experiments focus on the following research questions:

\begin{itemize}
    \item \textbf{RQ1:} How does ChiseLLM models perform on Spec-to-Chisel and Decompile-to-Chisel tasks compared to baseline models?
    \item \textbf{RQ2:} Can ChiseLLM models more effectively produce variability designs?
    \item \textbf{RQ3:} What is the impact of ChiseLLM- datasets on model performance?
\end{itemize}

\subsection{Experiment Setup}
\label{sec:experiment-setup}

\textbf{Benchmarks.} To the best of our knowledge, there are no publicly available evaluation datasets specifically for the Chisel. However, since Chisel can generate synthesizable Verilog code, we utilize widely-adopted Verilog evaluation datasets for our assessment. We use RTLLM v2.0~\cite{liuOpenLLMRTLOpenDataset} and VerilogEval-Human~\cite{pinckneyRevisitingVerilogEvalYear2025} datasets as benchmarks for evaluating syntactic/semantic correctness.

\textbf{Correctness Evaluation Method.} To measure the syntactic and semantic correctness of generated Chisel code, we employ the common Pass@k metric. We first use the Scala 2.13 compiler to compile the generated Chisel code, checking its syntactic correctness; then use Chisel3's \texttt{ChiselStage.emitSystemVerilog} API to convert the generated Chisel code to equivalent SystemVerilog code. In most cases, the SystemVerilog code compiled by this API is syntactically correct and synthesizable. Finally, we compile and simulate the code with testbench using commercial tools (necessary because the SystemVerilog syntax automatically generated by Chisel cannot be compiled by iVerilog) to check its functional correctness.

\textbf{Variability Evaluation Method.} To evaluate the variability of Chisel code, we adopt the mainstream LLM-as-a-Judge~\cite{zheng2023judgingllmasajudgemtbenchchatbot} method, which is considered close to human evaluation in code quality assessment~\cite{wangCanLLMsReplace2025}. To ensure consistency and fairness of evaluation results, we: (1) provide a publicly accessible evaluation standard to the Judge LLM, and (2) generate variants for each design specification in the evaluation datasets, and supplement these variants to the Judge LLM as a consistent baseline. All evaluation datasets are publicly accessible. The Judge LLM performs multiple evaluations on the same Chisel code. The final evaluation score is the average of all evaluation scores. Samples with high evaluation variance are not included in the evaluation. Similar to the Pass@k metric, we require the model to implement the same design requirement multiple times, and evaluate each implementation, with the average score of all implementations as the evaluation result.

\textbf{Baselines.} We select models with verified code generation performance as baselines, including: (1) Open-source non-reasoning models with general code generation capabilities, including Qwen2.5-\{7,32\}B-Coder-Instruct~\cite{hui2024qwen25coder}, Qwen2.5-72B-Instruct~\cite{qwen2025qwen25technicalreport}, and Llama-3.3-70B-Instruct~\cite{grattafiori2024llama3herdmodels}; (2) Open-source reasoning models with general code generation capabilities, including Deepseek-R1-Distill-Qwen-\{32,70\}B~\cite{deepseekaiDeepSeekR1IncentivizingReasoning2025} (excluding Deepseek-R1-Distill-Qwen-7B as it primarily focuses on solving mathematical problems); (3) Commercial models, including Deepseek-\{V3,R1\}~\cite{deepseekai2025deepseekv3technicalreport,deepseekaiDeepSeekR1IncentivizingReasoning2025} and GPT-4o~\cite{HelloGPT4o}. Other models including programming competition reasoning models~\cite{HuggingfaceOpenr1Fully} (specifically trained to programming in Python or C++) and models specifically fine-tuned for Verilog~\cite{zhaoCodeVEmpoweringLLMs2024,pei2024bettervcontrolledveriloggeneration} (Cannot generate correct Chisel code) are not considered as baselines to ensure fairness.


\subsection{Correctness Evaluation (RQ1)}
\label{sec:experiment-rq1}

Table~\ref{tab:rq1} shows the performance of various models on Spec-to-Chisel and Decompile-to-Chisel tasks on the VerilogEval and RTLLM datasets. We observe the following facts:

\textbf{General reasoning models do not consistently outperform their base models.} For instance, on the RTLLM dataset's Spec-to-Chisel task, Deepseek-R1-Distill-Qwen-32B achieves only 14.86\% pass@1, while its base model Qwen2.5-Coder-32B-Instruct reaches 26.50\%; similarly, Deepseek-R1-Distill-Llama-8B achieves 9.31\% pass@1 on the VerilogEval, lower than its base model Llama3.1-8B-Instruct. These observations indicates that there is not significant and consistent improvement for reasoning models over their base models.

\textbf{ChiseLLM models outperforms baseline models of similar size on all tasks, and even surpasses commercial models on some tasks.} Taking the RTLLM Decompile-to-Chisel task as an example, ChiseLLM-32B achieves 54.32\% pass@1, significantly higher than Qwen2.5-Coder-32B-Instruct's 13.00\%, and even exceeding GPT-4o's 44.50\%; ChiseLLM-7B achieves 50.47\% pass@1 on the VerilogEval dataset, nearly twice that of Qwen2.5-Coder-7B-Instruct (27.60\%). Notably, \textit{ChiseLLM-32B consistently ranks as the top or second-best performer across all metrics, demonstrating performance comparable to that of Deepseek-R1 despite having a significantly smaller parameter scale (32B vs 671B)}.

\subsection{Variability Evaluation (RQ2)}
\label{sec:experiment-rq2}

\input{tables/rq3}
\input{figures/rq2}

Figure~\ref{fig:rq2} shows the variability design capability of different models on the RTLLM dataset when performing Decompile-to-Chisel tasks. We observe the following facts:

\textbf{ChiseLLM models excel in variability design compared to baseline models and perform on par with commercial models.} ChiseLLM-32B's variability score is significantly higher than non-commercial baseline models such as Qwen-2.5-Coder-32B (3.79) and LLaMA-3.3-70B (2.39), and even exceeds commercial models like Deepseek-R1 (5.03) and Deepseek-V3 (4.84). This indicates that ChiseLLM-32B has acquired superior variability design capabilities through domain-specific fine-tuning.

\textbf{Reasoning models do not significantly dominate the variability design task.} Despite Deepseek-R1-Distill series models performing well on some code generation tasks, their performance in Chisel variability design is limited, with Deepseek-R1-Distill-32B and Deepseek-R1-Distill-70B achieving means of only 3.82 and 3.20, respectively. These values show limited gains over their corresponding base models Qwen-2.5-Coder-32B (3.79) and LLaMA-3.3-70B (2.39), indicating that general reasoning ability does not always positively correlate with domain-specific variability design capability.

\textbf{Better variability design capability is accompanied by greater uncertainty.} Although ChiseLLM-32B has the highest mean (5.19), it also has the largest variance; similarly, Deepseek-R1 and Deepseek-V3 also exhibit high means and variances. This suggests that when models attempt to implement more flexible designs, they may introduce more uncertainty. Smaller models like LLaMA-3.3-70B have lower variance (1.23) but also the weakest variability design capability (mean only 2.39). Overall, a higher mean indicates stronger variability design capability, but the diversity of designs may also bring greater performance fluctuations.

\subsection{Ablation Study (RQ3)}
\label{sec:experiment-rq3}

Table~\ref{tab:rq3} shows the results of our ablation study. To demonstrate the impact of ChiseLLM datasets on model performance, we fine-tuned ChiseLLM-32B model separately using these two datasets, resulting in ChiseLLM-32B-Completion and ChiseLLM-32B-Decompile models, and evaluated their performance. We observe the following facts:

\input{figures/case-study-2}

\textbf{Both datasets individually improve model performance.} ChiseLLM-\{Completion,Decompile\} datasets each have a positive impact on model performance. For example, on the VerilogEval Decompile-to-Chisel task, ChiseLLM-32B-Completion improves pass@1 by 2.88 percentage points, while ChiseLLM-32B-Decompile improves it by 14.78 percentage points; on the RTLLM Decompile-to-Chisel task, ChiseLLM-32B-Completion improves pass@1 by 23.33 percentage points, and ChiseLLM-32B-Decompile improves it by 37.56 percentage points.

\textbf{The combination of both datasets produces significant synergistic effects.} Across all tasks, the ChiseLLM-32B model trained with both datasets outperforms models trained with either dataset alone. For example, on the VerilogEval Decompile-to-Chisel task, ChiseLLM-32B achieves 56.41\% pass@1, higher than both ChiseLLM-32B-Decompile's 55.97\% and ChiseLLM-32B-Completion's 44.07\%; on the RTLLM Spec-to-Chisel task, ChiseLLM-32B achieves a syntax correctness rate of 64.05\%, higher than both ChiseLLM-32B-Decompile's 60.41\% and ChiseLLM-32B-Completion's 43.08\%.

%% file: tables/rq1.tex
\begin{table*}[t!]
    \begin{threeparttable}
        \caption{Model performance on VerilogEval-Human and RTLLM-V2.0 datasets.}
        \label{tab:rq1}
        \centering
        \small
        \begin{tabular}{l@{\hspace{4pt}} ccc@{\hspace{6pt}} ccc@{\hspace{6pt}} ccc@{\hspace{6pt}} ccc}
            \toprule
            \multirow{4}{*}{Model} & \multicolumn{6}{c}{\textbf{VerilogEval-Human}} & \multicolumn{6}{c}{\textbf{RTLLM-V2.0}}                                                                                                                                                                                                                                                                                                                                                                                         \\
            \cmidrule(lr){2-7} \cmidrule(lr){8-13}
                                   & \multicolumn{3}{c}{Spec-to-Chisel}             & \multicolumn{3}{c}{Decompile-to-Chisel} & \multicolumn{3}{c}{Spec-to-Chisel}  & \multicolumn{3}{c}{Decompile-to-Chisel}                                                                                                                                                                                                                                                                                                         \\
            \cmidrule(lr){2-4} \cmidrule(lr){5-7} \cmidrule(lr){8-10} \cmidrule(lr){11-13}
                                   & P@1 \tnote{1}                                  & P@5 \tnote{1}                           & Syn(\%) \tnote{1}                   & P@1                                     & P@5                                & Syn(\%)                            & P@1                                & P@5                                & Syn(\%)                            & P@1                                & P@5                                & Syn(\%)                            \\
            \midrule
            \multicolumn{13}{c}{\textbf{7B+ Scale Models}}                                                                                                                                                                                                                                                                                                                                                                                                                                                            \\
            \midrule
            Llama3.1-8B \tnote{2}  & 4.33                                           & 9.90                                    & 9.02                                & 5.43                                    & 12.29                              & 11.15                              & -                                  & -                                  & -                                  & -                                  & -                                  & -                                  \\
            Qwen2.5-Coder-7B       & 21.94                                          & 31.87                                   & 37.08                               & 27.60                                   & 34.58                              & 43.23                              & 7.73                               & 18.00                              & 15.56                              & 8.08                               & 11.44                              & 16.55                              \\
            *Dpsk-R1-8B \tnote{3}  & 9.31                                           & 15.44                                   & 16.01                               & 10.05                                   & 16.15                              & 12.03                              & -                                  & -                                  & -                                  & -                                  & -                                  & -                                  \\
            *ChiseLLM-7B           & \textbf{29.41}                                 & \textbf{47.08}                          & \textbf{58.82}                      & \textbf{50.47}                          & \textbf{70.99}                     & \textbf{59.19}                     & \textbf{22.00}                     & \textbf{37.46}                     & \textbf{41.78}                     & \textbf{33.29}                     & \textbf{51.11}                     & \textbf{45.42}                     \\
            \midrule
            \multicolumn{13}{c}{\textbf{32B+ Scale Models}}                                                                                                                                                                                                                                                                                                                                                                                                                                                           \\
            \midrule
            Qwen2.5-Coder-32B      & 41.02                                          & 53.85                                   & 73.47                               & 41.19                                   & 48.96                              & 53.93                              & 26.50                              & 38.39                              & 48.00                              & 13.00                              & 21.75                              & 28.25                              \\
            Qwen2.5-72B            & 39.74                                          & 49.30                                   & 61.31                               & 40.54                                   & 47.32                              & 59.30                              & 22.75                              & 31.79                              & 43.75                              & 20.25                              & 29.14                              & 30.00                              \\
            Llama-3.3-70B          & 38.14                                          & 44.90                                   & 65.97                               & 38.38                                   & 46.96                              & 48.00                              & 22.33                              & 31.27                              & 47.26                              & 20.00                              & 27.18                              & 43.00                              \\
            *Dpsk-R1-32B           & 38.50                                          & 54.58                                   & 52.19                               & 45.03                                   & 63.02                              & 53.17                              & 14.86                              & 27.00                              & 26.21                              & 13.11                              & 26.41                              & 19.65                              \\
            *Dpsk-R1-70B           & 36.62                                          & 52.28                                   & 51.72                               & 37.50                                   & 55.05                              & 45.59                              & 16.51                              & 29.10                              & 26.27                              & 15.00                              & 30.46                              & 22.81                              \\
            *ChiseLLM-32B          & \cellcolor{blue!60}\textbf{51.43}\tnote{4}     & \cellcolor{blue!60}\textbf{68.29}       & \cellcolor{blue!60}\textbf{76.45}   & \cellcolor{green!60}\textbf{56.41}      & \cellcolor{green!60}\textbf{72.00} & \cellcolor{blue!60}\textbf{64.71}  & \cellcolor{blue!60}\textbf{42.32}  & \cellcolor{blue!60}\textbf{57.37}  & \cellcolor{blue!60}\textbf{64.05}  & \cellcolor{blue!60}\textbf{54.32}  & \cellcolor{green!60}\textbf{70.56} & \cellcolor{green!60}\textbf{62.15} \\
            \midrule
            \multicolumn{13}{c}{\textbf{Commercial Models}}                                                                                                                                                                                                                                                                                                                                                                                                                                                           \\
            \midrule
            Dpsk-V3-671B           & 50.16                                          & 63.44                                   & 76.37                               & \cellcolor{blue!60}\textbf{54.57}       & 63.19                              & \cellcolor{green!60}\textbf{66.19} & 42.00                              & 52.32                              & \cellcolor{green!60}\textbf{73.25} & 49.25                              & 57.39                              & 58.00                              \\
            GPT-4o                 & 42.04                                          & 60.16                                   & 69.76                               & 42.39                                   & 65.75                              & 53.77                              & 34.50                              & 48.11                              & 60.75                              & 44.50                              & 64.00                              & 55.00                              \\
            *Dpsk-R1-671B          & \cellcolor{green!60}\textbf{62.74}\tnote{4}    & \cellcolor{green!60}\textbf{76.05}      & \cellcolor{green!60} \textbf{82.85} & 53.45                                   & \cellcolor{blue!60}\textbf{71.50}  & 59.13                              & \cellcolor{green!60}\textbf{46.75} & \cellcolor{green!60}\textbf{61.07} & 63.50                              & \cellcolor{green!60}\textbf{55.25} & \cellcolor{blue!60}\textbf{69.25}  & \cellcolor{blue!60}\textbf{61.75}  \\
            \bottomrule
        \end{tabular}
        \smallskip
        \scriptsize
        \begin{tablenotes}
            \RaggedRight
            \item[1] Syn(\%) represents the proportion of syntactically correct samples. P@k is the abbreviation of Pass@k metric.
            \item[2] Some 7B+ scale models lack results due to insufficient correct samples for statistical significance.
            \item[3] The models marked with asterisks (*) are reasoning models.
            \item[4] \colorbox{green!60}{Green cells} indicate the best performance, while \colorbox{blue!60}{blue cells} indicate the second-best performance.

        \end{tablenotes}
    \end{threeparttable}
\end{table*}

%% file: tables/rq3.tex
\begin{table*}[t!]
    \begin{threeparttable}
        \caption{ChiseLLM model performance when training on different datasets.}
        \label{tab:rq3}
        \centering
        \small
        \begin{tabular}{l@{\hspace{4pt}} ccc@{\hspace{6pt}} ccc}
            \toprule
            \multirow{3}{*}{Model}     & \multicolumn{3}{c}{\textbf{VerilogEval-Human}} & \multicolumn{3}{c}{\textbf{RTLLM-V2.0}}                                                                                                                                                       \\
            \cmidrule(lr){2-4} \cmidrule(lr){5-7}
                                       & P@1                                            & P@5                                     & Syntax\%                           & P@1                                 & P@5                                 & Syntax\%                           \\
            \midrule
            \multicolumn{7}{c}{\textbf{Spec-to-Chisel}}                                                                                                                                                                                                                                 \\
            \midrule
            Qwen2.5-Coder-32B-Inst     & 41.03 (0.0)                                    & 58.79 (0.0)                             & 73.47 (0.0)                        & 26.50 (0.0)                         & 42.82 (0.0)                         & 48.00 (0.0)                        \\
            ChiseLLM-32B-Comp\tnote{2} & 44.09 (\cellcolor{green!25}+3.06)\tnote{1}     & 69.31 (\cellcolor{green!45}+10.52)      & 66.42 (\cellcolor{red!45}-7.05)    & 25.64 (\cellcolor{red!30}-0.86)     & 48.18 (\cellcolor{green!30}+5.36)   & 43.08 (\cellcolor{red!35}-4.92)    \\
            ChiseLLM-32B-De\tnote{2}   & 49.51 (\cellcolor{green!40}+8.48)              & 71.41 (\cellcolor{green!55}+12.62)      & 76.80 (\cellcolor{green!25}+3.33)  & 37.18 (\cellcolor{green!45}+10.68)  & 55.90 (\cellcolor{green!55}+13.08)  & 60.41 (\cellcolor{green!55}+12.41) \\
            ChiseLLM-32B               & 51.43 (\cellcolor{green!45}+10.4)              & 72.78 (\cellcolor{green!60}+13.99)      & 76.45 (\cellcolor{green!25}+2.98)  & 42.32 (\cellcolor{green!65}+15.82)  & 61.43 (\cellcolor{green!75}+18.61)  & 64.05 (\cellcolor{green!65}+16.05) \\
            \midrule
            \multicolumn{7}{c}{\textbf{Decompile-to-Chisel}}                                                                                                                                                                                                                            \\
            \midrule
            Qwen2.5-Coder-32B-Inst     & 41.19 (0.0)                                    & 51.59 (0.0)                             & 53.93 (0.0)                        & 13.00 (0.0)                         & 26.39 (0.0)                         & 28.25 (0.0)                        \\
            ChiseLLM-32B-Comp          & 44.07 (\cellcolor{green!25}+2.88)              & 71.41 (\cellcolor{green!75}+19.82)      & 54.25 (\cellcolor{green!15}+0.32)  & 36.33 (\cellcolor{green!85}+23.33)  & 66.85 (\cellcolor{green!100}+40.46) & 43.23 (\cellcolor{green!60}+14.98) \\
            ChiseLLM-32B-De            & 55.97 (\cellcolor{green!60}+14.78)             & 74.92 (\cellcolor{green!85}+23.33)      & 60.93 (\cellcolor{green!35}+7.0)   & 50.56 (\cellcolor{green!95}+37.56)  & 74.26 (\cellcolor{green!100}+47.87) & 55.83 (\cellcolor{green!90}+27.58) \\
            ChiseLLM-32B               & 56.41 (\cellcolor{green!60}+15.22)             & 77.67 (\cellcolor{green!90}+26.08)      & 64.71 (\cellcolor{green!45}+10.78) & 54.32 (\cellcolor{green!100}+41.32) & 76.15 (\cellcolor{green!100}+49.76) & 62.15 (\cellcolor{green!95}+33.90) \\
            \bottomrule
        \end{tabular}
        \smallskip
        \scriptsize
        \begin{tablenotes}
            \RaggedRight
            \item[1] The values in brackets represent the performance improvement compared to the baseline model Qwen2.5-Coder-32B-Instruct.
            \item[2] ChiseLLM-Completion and ChiseLLM-Decompile are models trained on single dataset, while ChiseLLM combines both datasets.
        \end{tablenotes}
    \end{threeparttable}
\end{table*}

%% file: figures/rq2.tex
\begin{figure}
    \begin{small}
        \centering
        \includegraphics[width=0.9\linewidth]{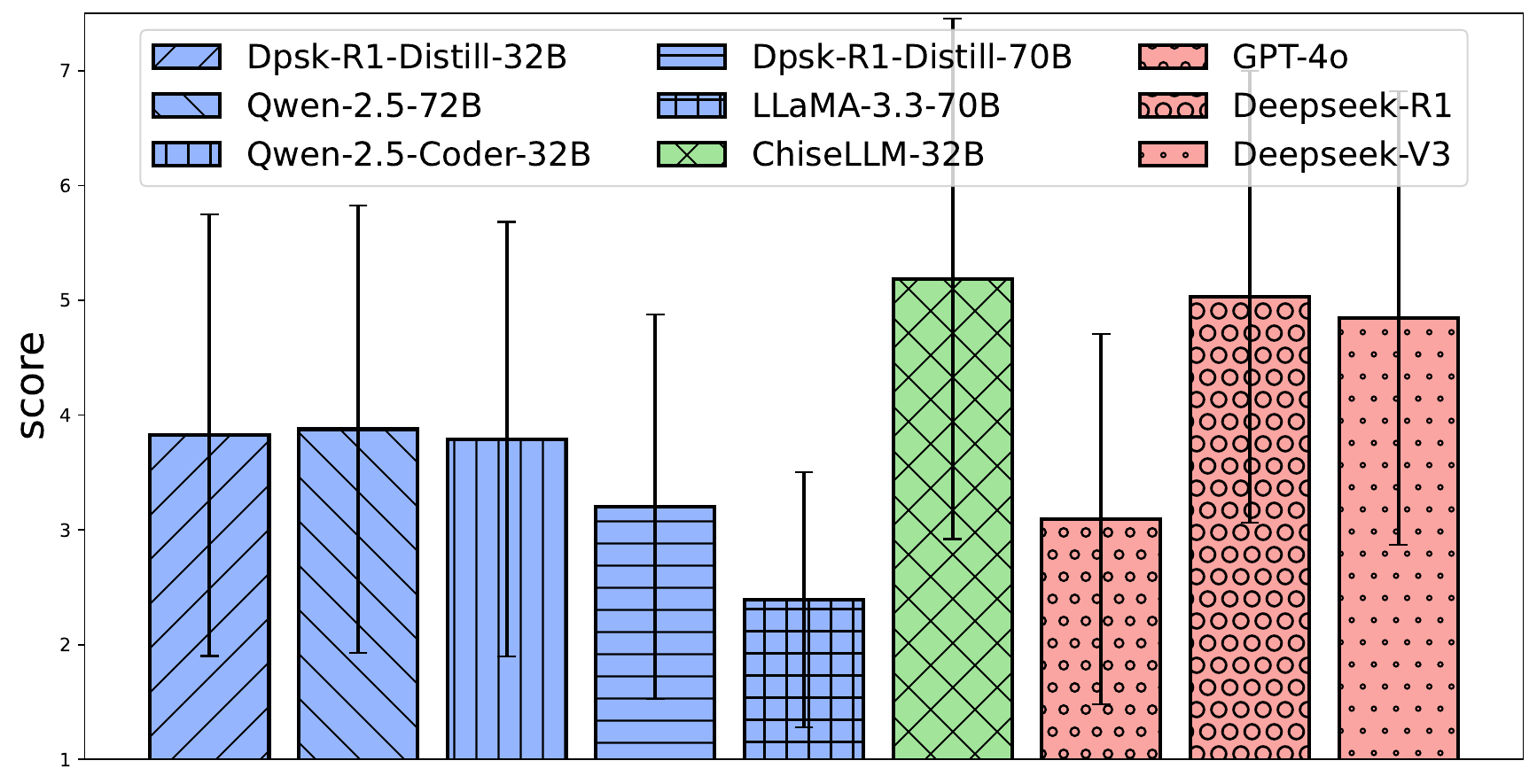}
        \caption{Variability design capability of different models on the RTLLM dataset when performing Decompile-to-Chisel tasks. The blue bar represents the open-source reasoning and non-reasoning baseline models. The green bar represents the ChiseLLM-32B model. The red bars represent the commercial models. Each bar is distinctively hatched with different patterns.}
        \label{fig:rq2}
    \end{small}
\end{figure}

%% file: figures/case-study-2.tex
\begin{figure*}[t!]
    \centering
    \includegraphics[width=0.80\textwidth]{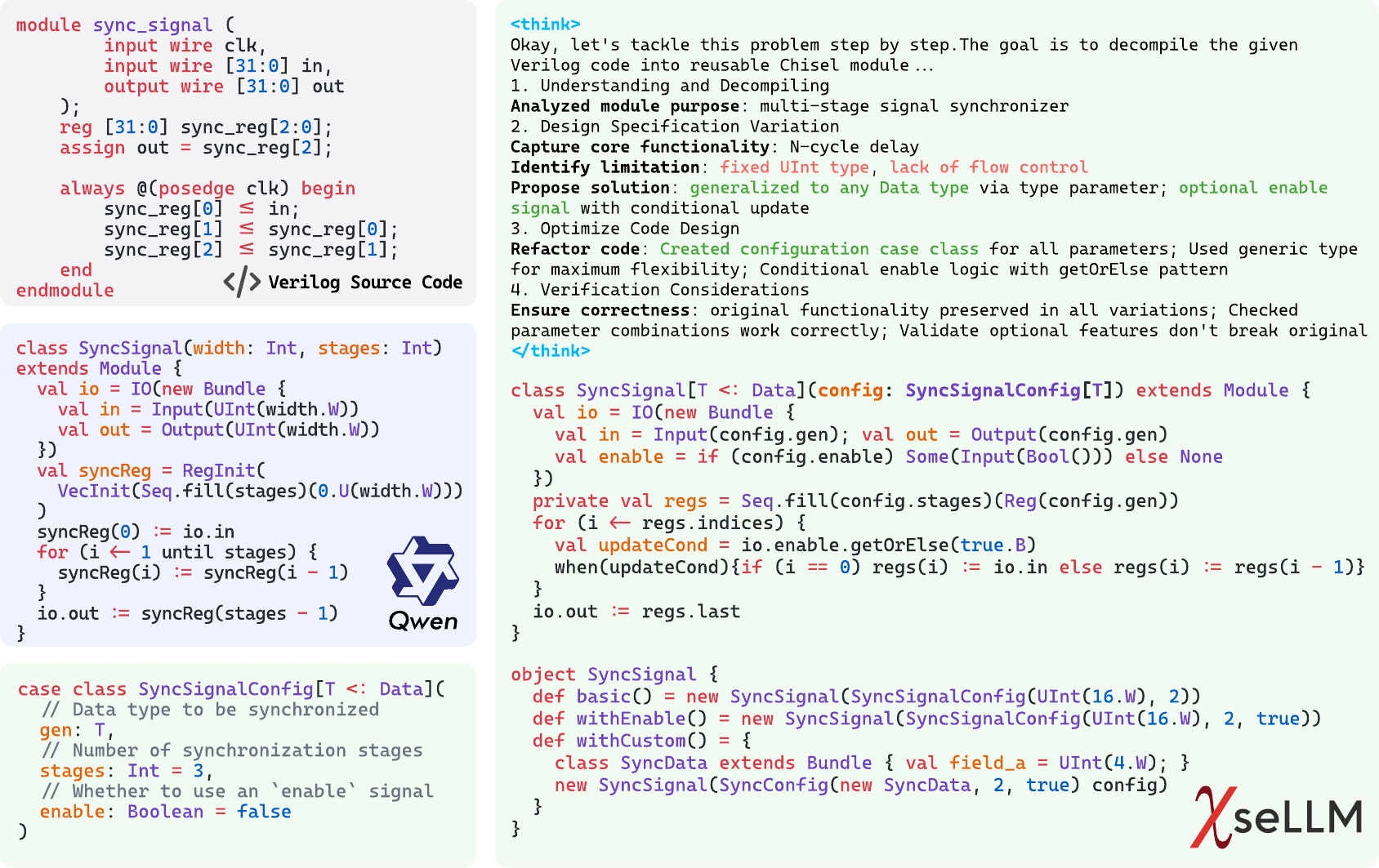}
    \caption{An actual example of ChiseLLM-32B and Qwen2.5-32B-Coder-Instruct decompiling Verilog source code. The gray part represents the Verilog source code, the blue part represents the Chisel module generated by the Qwen model, and the green part represents the Chisel module generated by the ChiseLLM model. The content within the \texttt{<think>} tags represents the thinking process of the ChiseLLM model during decompilation, which, due to its length, is summarized as a numbered list in the figure. The ChiseLLM model follows a structured reasoning process during decompilation, ultimately producing a Chisel module with higher variability and functional extensibility.}
    \label{fig:case-study-2}
\end{figure*}

%% file: sections/case-study.tex
\section{Case Study}
\label{sec:case-study}

\input{figures/case-study-1}
In this section, we present two case studies to demonstrate and compare the design variability capabilities of the fine-tuned ChiseLLM-32B model. The purpose of this section is to complement the results presented in Section~\ref{sec:experiment-rq2} by providing intuitive examples that illustrate both the thinking process during design and the distinctive characteristics of the final designs produced by the ChiseLLM models.

Figure~\ref{fig:case-study-1} illustrates an actual example of ChiseLLM-32B decompiling Verilog source code. In this example, the original Verilog code implements a simple shift module with fixed data width and shift patterns. During the decompilation process, ChiseLLM considered both configurational and structural design variants by introducing the \texttt{width} configuration parameter and the abstract \texttt{ShifterBase} class. These enhancements transformed the module into a highly variable and extensible Chisel implementation with configurable data (variable width) and structure (customizable shift behavior). Although the code shown in this example is relatively simple due to space constraints, the ChiseLLM models can apply the same logical approach to process more complex instances in practical applications.

Figure~\ref{fig:case-study-2} presents a comparative example where both ChiseLLM and Qwen2.5-32B-Coder-Instruct decompile the same Verilog source code. In this case, the Verilog module named \texttt{sync\_signal} is designed to delay input data by multiple clock cycles before outputting it. The Chisel implementation generated by the Qwen model includes some optimizations but only considers configurational variability by parameterizing the data width and synchronization cycles. In contrast, ChiseLLM follows a structured reasoning process, first analyzing the core functionality of the module, then identifying key deficiencies in the current implementation—fixed data structure and lack of flow control—and addressing these issues by incorporating appropriate Chisel language features. The resulting Chisel module from ChiseLLM can process arbitrary complex types of data input and control data flow transmission through enable signals, thus exhibiting greater variability and functional extensibility.

%% file: figures/case-study-1.tex
\begin{figure}[ht!]
    \centering
    \includegraphics[width=0.73\linewidth]{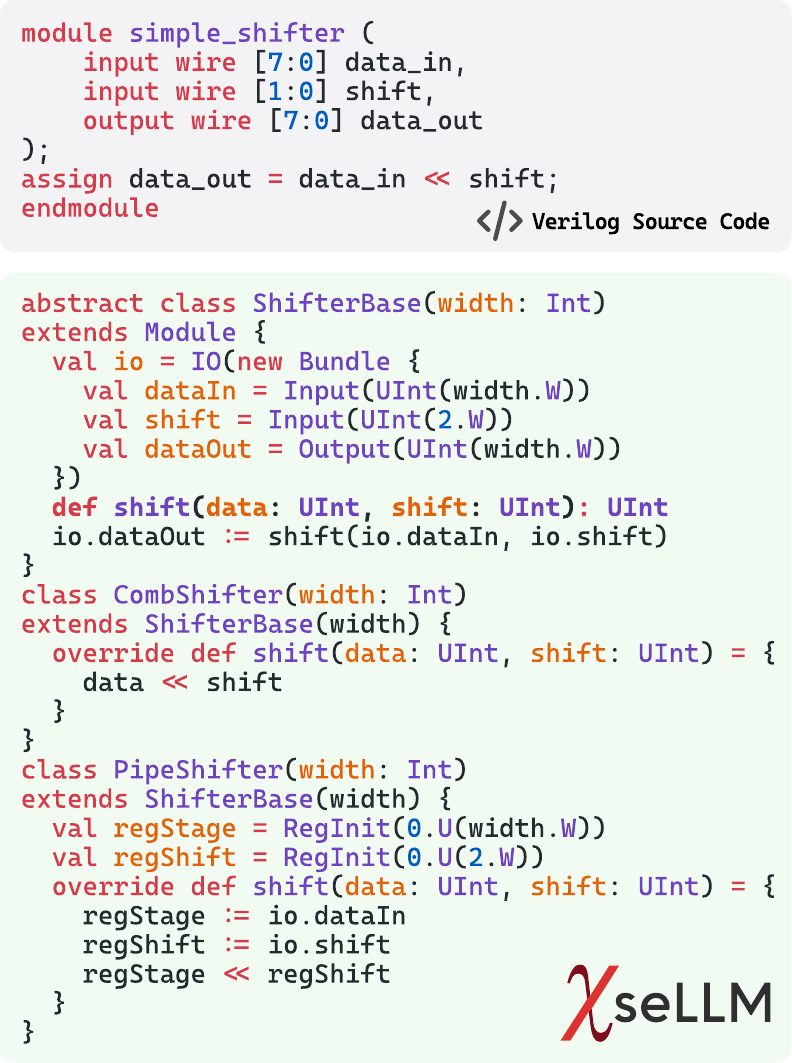}
    \caption{An actual example of ChiseLLM-32B decompiling Verilog source code. The gray part represents the Verilog source code, and the green part represents the Chisel module generated by ChiseLLM. While ensuring all functionalities of the Verilog source code are covered, the module generated by ChiseLLM introduces advanced language features, resulting in higher variability and functional extensibility.}
    \label{fig:case-study-1}
\end{figure}

%% file: sections/conclusion.tex
\section{Conclusion}
\label{sec:conclusion}

In this paper, we presented ChiseLLM, comprising a high-quality reasoning dataset and an inference model. We collected training data from public resources and enhanced LLM performance in HCL generation through prompt-guided distillation. Our work provides two promising perspectives for future AHDM: First, high-quality domain adaptation training corpora can significantly improve LLM performance in HCL generation. Second, with additional guidance, LLMs can master the "thinking patterns" necessary to address the complexities in hardware design. As an initial exploration in LLM-assisted HCL-Based AHDM, we hope to draw more researchers' attention to this field and advance the application of LLMs in hardware design.